\documentclass{article}

\usepackage{microtype}
\usepackage{graphicx}
\usepackage{subcaption}
\usepackage{booktabs} 
\usepackage{booktabs}  
\usepackage{multirow}  
\usepackage{graphicx}  
\usepackage{enumitem}
\usepackage{hyperref}
\usepackage{tcolorbox}
\tcbuselibrary{breakable, skins}
\usepackage{fvextra} 
\usepackage{float}

\usepackage[preprint]{icml2026}

\usepackage{amsmath}
\usepackage{amssymb}
\usepackage{mathtools}
\usepackage{amsthm}

\usepackage[capitalize,noabbrev]{cleveref}

\theoremstyle{plain}

\theoremstyle{definition}

\theoremstyle{remark}

\usepackage[textsize=tiny]{todonotes}

\icmltitlerunning{Hindsight Credit Assignment for Long-Horizon LLM Agents}

\begin{document}

\twocolumn[
  \icmltitle{Hindsight Credit Assignment for Long-Horizon LLM Agents}



  \icmlsetsymbol{equal}{*}

  \begin{icmlauthorlist}
    \icmlauthor{Hui-Ze Tan*}{lab,nju}
    \icmlauthor{Xiao-Wen Yang*}{lab,nju}
    \icmlauthor{Hao Chen*}{tencent}
    \icmlauthor{Jie-Jing Shao}{lab}
    \icmlauthor{Yi Wen}{tencent,cityu}
    \icmlauthor{Yuteng Shen}{tencent}
    \icmlauthor{Weihong Luo}{tencent}
    \icmlauthor{Xiku Du}{tencent}
    \icmlauthor{Lan-Zhe Guo}{lab,njusz}
    \icmlauthor{Yu-Feng Li}{lab,nju}
  \end{icmlauthorlist}

  \icmlaffiliation{lab}{State Key Laboratory of Novel Software Technology, Nanjing University, China}
  \icmlaffiliation{nju}{School of Artificial Intelligence, Nanjing University, China}
  \icmlaffiliation{njusz}{School of Intelligence Science and Technology, Nanjing University, China}
  \icmlaffiliation{tencent}{FiT, Tencent, Shenzhen, China}
  \icmlaffiliation{cityu}{School of Data Science, City University of Hong Kong, Hong Kong, China}

  \icmlcorrespondingauthor{Weihong Luo}{lobbyluo@tencent.com}
  \icmlcorrespondingauthor{Yu-Feng Li}{liyf@nju.edu.cn}

  \icmlkeywords{Machine Learning, ICML}

  \vskip 0.3in
]



\printAffiliationsAndNotice{}  

\begin{abstract}
Large Language Model (LLM) agents often face significant credit assignment challenges in long-horizon, multi-step tasks due to sparse rewards. Existing value-free methods, such as Group Relative Policy Optimization (GRPO), encounter two fundamental bottlenecks: inaccurate step-level $Q$-value estimation and misaligned value baselines for intermediate states. To address these limitations, we introduce \textbf{HCAPO}, the first framework to integrate hindsight credit assignment into LLM agents. HCAPO leverages the LLM itself as a post-hoc critic to refine step-level $Q$-values through hindsight reasoning. Furthermore, HCAPO's multi-scale advantage mechanism effectively supplements the inaccurate value baselines at critical decision states. Evaluations across three challenging benchmarks, including WebShop and ALFWorld, demonstrate that HCAPO consistently outperforms state-of-the-art RL methods. Notably, HCAPO achieves a 7.7\% improvement in success rate on WebShop and a 13.8\%  on ALFWorld over GRPO using the Qwen2.5-7B-Instruct model. These results indicate that HCAPO significantly enhances exploration efficiency, promotes concise decision-making, and ensures scalability in complex, long-horizon tasks.
\end{abstract}

\section{Introduction}
\begin{figure}[t]
    \centering
    \includegraphics[width=\linewidth]{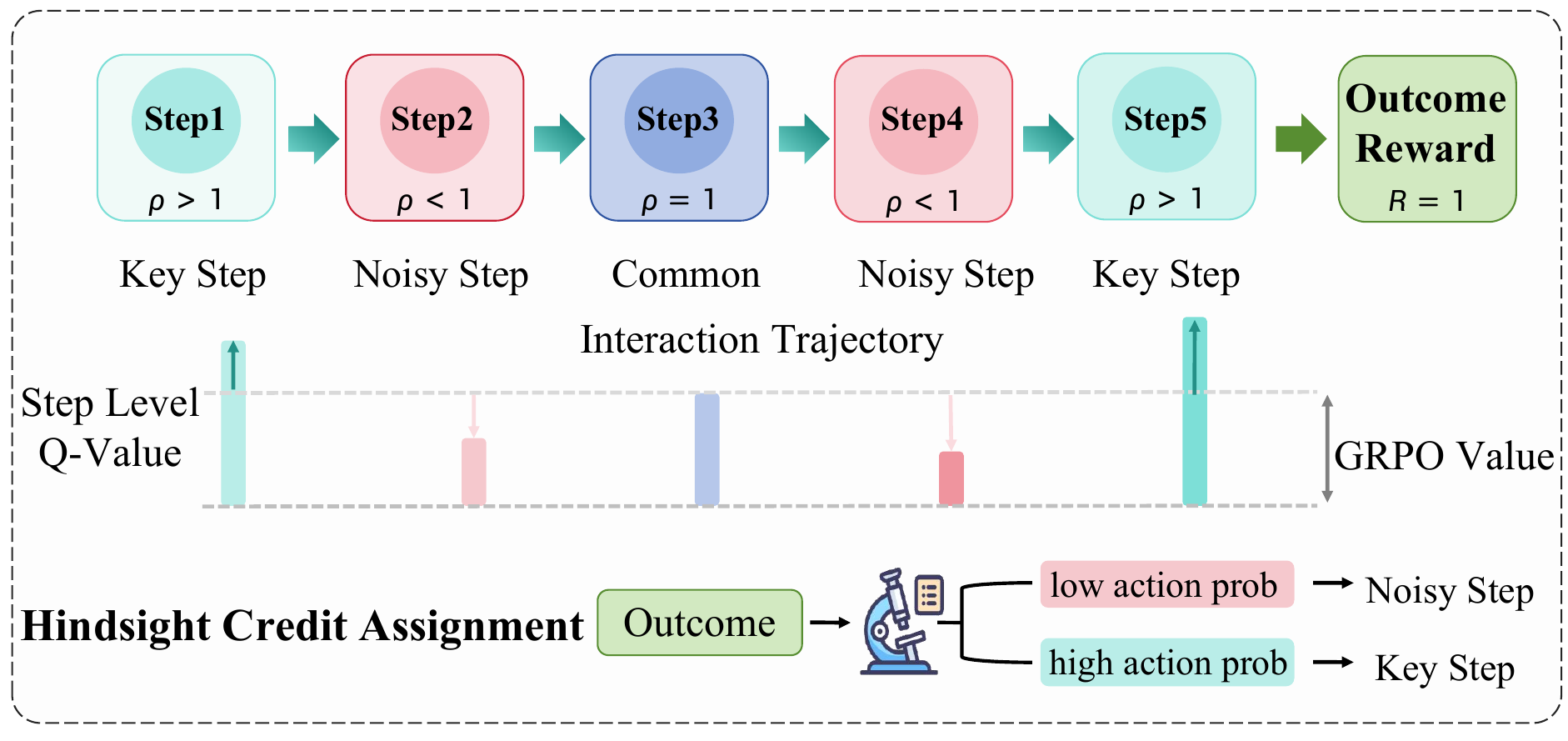}
    \caption{From trajectory-level to step-level: hindsight credit assignment for long-horizon agents. $\rho$ is the hindsight ratio.}
    \label{fig:intro}
\end{figure}
In recent years, Large Language Model (LLM)-based autonomous agents have demonstrated remarkable advancements in reasoning and decision-making within open environments~\citep{webagent,searchr1,deepseek-v3,codeagent}. These agents exhibit significant potential in addressing long-horizon planning tasks, including embodied planning~\citep{alfworld}, web navigation~\citep{webshop}, deep search, and multi-step travel planning~\citep{xie2024travelplanner,shao2024chinatravel}. 

Despite these advancements, a fundamental bottleneck persists in the application of reinforcement learning (RL)~\citep{reinforcement-learning,sutton1999policy} for agent optimization: the inherent sparsity of outcome-based rewards. Since most tasks provide only a scalar reward upon reaching a terminal state, intermediate actions within the decision-making process lack timely or granular feedback. This leads to a critical \textit{credit assignment} problem, where it becomes difficult to accurately attribute a sparse terminal reward to the specific, pivotal decisions that led to the final outcome. This challenge is further aggravated by the extended reasoning chains and vast action spaces of LLMs. 

To be more specific, we identify two fundamental bottlenecks in applying current value-free methods like GRPO~\citep{deepseekmath,deepseek-r1} to agent tasks. First, the \textit{inaccuracy of step-level Q-value estimation}: since these methods rely on a single Monte Carlo sample (the terminal reward) for the entire trajectory, they fail to discern the specific contribution of individual actions. Second, the \textit{misalignment of the value baseline}: GRPO typically utilizes the mean reward from the initial state as a universal baseline, failing to account for the evolving state values as the agent progresses through a long sequence of interactions.

To mitigate this challenge, existing research has sought to construct dense reward signals by incorporating task-specific priors or leveraging external models.   For instance, GIGPO~\citep{gigpo} utilizes anchor states to categorize trajectories, thereby optimizing advantage estimation, while EMPG~\citep{empg} formulates intrinsic, action-level rewards based on dynamic entropy.   Furthermore, several approaches have integrated Process Reward Models (PRMs) to provide fine-grained, step-by-step feedback~\citep{prm,agentprm}.   However, the development of PRMs relies heavily on costly human annotations and is susceptible to noise, which limits their generalization capabilities in out-of-distribution scenarios.  More crucially, current methods predominantly focus on a unidirectional forward process (from initial to goal states), while overlooking the retrospective causal link between specific intermediate actions and the final outcome.  

In classical RL, hindsight methods offer a promising alternative by leveraging information available after an episode concludes, especially for long-horizon tasks~\citep{hca,her}. The intuition is powerful: once we know a trajectory succeeded, we can look back and ask, ``Given this successful outcome, how necessary was each action?" If an action aligns strongly with the path to success, it deserves amplified credit; if it appears irrelevant or suboptimal in hindsight, its credit should be suppressed. This approach helps uncover causal relationships between intermediate decisions and final outcomes. However, effectively implementing this intuition within the unique constraints of LLM agents, where the action space is combinatorial natural language and the policy is a generative model, remains an open challenge.

In this paper, we introduce Hindsight Credit Assignment Policy Optimization (HCAPO), a novel, value-free framework designed to address sparse-reward training for long-horizon LLM agents (see Figure~\ref{fig:intro}). Our key contributions are summarized as follows:

\begin{itemize}[leftmargin=1em,nosep]
    \item \textbf{A Principled Hindsight Framework:} We introduce HCAPO, the first framework to integrate hindsight credit assignment into LLM agents. We propose Generative Verification, which leverages the LLM itself as a post-hoc critic to evaluate instrumental actions by conditioning on successful outcomes. We further introduce a self-normalized importance ratio estimation that bypasses the need for external models. By refining step-level Q-values, HCAPO effectively mitigates the issues of credit assignment in current value-free methods like GRPO.
    
    \item \textbf{Theoretical Insights into Multi-Scale Advantages:} We provide a formal analysis for HCAPO's composite advantage mechanism. We demonstrate that HCAPO addresses two fundamental limitations of standard group optimization: the coarse estimation of step-level Q-values and the misalignment of value estimation for intermediate states. Our analysis shows that by refining Q-values and employing multi-scale advantage integration, HCAPO provides an accurate value estimate specifically at critical bottleneck nodes, while leveraging robust trajectory-level signals to maintain global training stability.
    

    \item \textbf{Empirical Superiority and Scalability:} Evaluations across ALFWorld, WebShop, and Search-augmented QA show that HCAPO consistently outperforms state-of-the-art RL methods. We benchmark HCAPO against the strong value-free baseline GRPO on both ALFWorld and WebShop.On WebShop, HCAPO raises the 7B-model success rate from 66.1\% $\to$ 73.8\% ($+$7.7\%).On ALFWorld, the gain is larger: 77.6\% $\to$ 91.4\% ($+$13.8\%), and with temporal smoothing the same model reaches 96.9\%, near-perfect.
\end{itemize}

\section{Related Work}
\label{sec:related_work}

\textbf{LLMs as Autonomous Agents.} 
LLMs have demonstrated significant potential as autonomous agents capable of reasoning, planning, and interacting with diverse environments \citep{react,reflexion,toolformer,ufo}. By leveraging their vast world knowledge, these agents can solve complex, multi-step tasks such as web navigation \citep{webshop} and embodied planning \citep{alfworld}. However, as the task horizon extends, agents often struggle with error accumulation and the lack of intermediate guidance, necessitating more effective optimization strategies beyond simple prompting.

\textbf{Reinforcement Learning for LLM Agents.} 
RL has become a pivotal paradigm for aligning LLMs with complex objectives~\citep{ouyang2022training,ziegler2019fine,stiennon2020learning}. While PPO~\citep{ppo} is a standard, its reliance on a learned Critic incurs significant memory overhead. Consequently, value-free methods like RLOO~\citep{rloo}, GRPO~\citep{deepseekmath} and other methods~\citep{dapo,liu2025understanding,lin2025cppo} have emerged to estimate advantages via group statistics. However, these methods primarily focus on trajectory-level feedback, which is often too coarse for long-horizon tasks where success hinges on pivotal actions.What's more, global baselines from initial states do not adapt to intermediate states, providing poor  signals.

\textbf{Reward Shaping and Process Supervision.} 
To address the sparse reward challenge, various methods have been proposed to tackle the credit assignment problem in LLM-based RL \citep{zhang2025rlvmr, liu2025agentic,li2025salt,dong2025agentic, zhou2024archer}. Specifically, Process Reward Models (PRMs) \citep{prm} provide step-level supervision but require expensive human annotations. Alternatively, intrinsic reward mechanisms such as EMPG \citep{empg} utilize dynamic entropy for exploration. More recently, GiGPO \citep{gigpo} introduced state-based anchors to categorize trajectories and refine advantages. Unlike these methods, HCAPO requires no manual anchor rules or external models, instead leveraging the LLM's intrinsic reasoning for  credit assignment.

\section{Preliminaries}
\subsection{RL Framework for LLM Agent Tasks}
We formalize interactive decision-making tasks as a Partially Observable Markov Decision Process (POMDP)~\citep{spaan2012partially}. At each time step $t$, the agent receives an observation $o_t$ (e.g., HTML source code), which, combined with the action history, constitutes the current state $s_t$. The agent then generates an action $a_t$ (e.g., clicking a button or issuing a search query) according to a policy $\pi_\theta(a_t | s_t)$. This interaction results in a trajectory of length $T$, denoted as $\tau = (s_1, a_1, \dots, s_T, a_T)$. 

These tasks are typically characterized by sparse rewards. The environment provides a scalar reward $R(\tau)$ only at the end of the task ($t=T$) based on the completion status
\begin{equation}
J(\pi_\theta) = \mathbb{E}_{\tau \sim \pi_\theta} [R(\tau)]
\end{equation}

\subsection{Value-Free Group Policy Optimization}
Direct optimization of the above objective typically relies on policy gradient methods. The standard gradient estimate takes the form:
\begin{equation}
\nabla_\theta J(\pi_\theta) = \mathbb{E}_{\tau \sim \pi_\theta} \left[ \sum_{t=0}^{T} A_t \nabla_\theta \log \pi_\theta(a_t|s_t) \right]
\end{equation}
where $A_t$ is the advantage function measuring the relative quality of action $a_t$. In traditional RL, $A_t$ is often estimated using a learned value network (Critic) $V(s)$ to reduce variance~\citep{dqn,gae}. However, in the context of Large Language Models (LLMs), training a Critic of comparable size to the Policy incurs significant memory overhead and training instability. Moreover, value estimation suffers from high bias in long-horizon, sparse-reward settings. Consequently, value-free methods~\citep{buy-4-reinforce-samples,dpo,li2023remax} have emerged as an efficient paradigm.

\textbf{Group Relative Policy Optimization (GRPO)} \citep{deepseekmath} exemplifies this paradigm. Instead of training a Critic, GRPO samples a group of $G$ trajectories $\{\tau_1, \dots, \tau_G\}$ for each input query using the current policy. It utilizes group statistics as a baseline to compute the advantage:
\begin{equation}
A_i^{\text{GRPO}} = \frac{R(\tau_i) - \mu_R}{\sigma_R}
\end{equation}
where $\mu_R$ and $\sigma_R$ are the group statistics of outcome rewards.This approach effectively reduces gradient variance through intra-group comparison without requiring additional value network parameters.

However, GRPO and related value-free methods encounter fundamental limitations for credit assignments in long-horizon agent tasks. First, because the advantage $A_i^{\text{GRPO}}$ is derived solely from the terminal return $R(\tau_i)$ of a complete trajectory, these methods lack the granularity for accurate step-level Q-value estimation, failing to distinguish critical actions from irrelevant ones. Second, the reliance on a global baseline from the initial state results in a misalignment with the evolving state values during extended interactions. Consequently, establishing precise step-level credit assignment remains a pivotal challenge for optimizing LLM agents in environments with sparse rewards.

\subsection{Hindsight Credit Assignment}

\label{sec:prelim_hca}

HCA ~\citep{hca} addresses this limitation by leveraging future outcome information to disentangle the contribution of individual steps. Its core idea is to introduce a hypothetical \textit{hindsight distribution} conditioned on the realized outcome.

Formally, let $\pi(a_t|s_t)$ denote the behavior policy used during sampling. We define a future-state-conditional distribution $h(a_t|s_t, s_k)$, which represents the probability of taking action $a_t$ at state $s_t$, \textit{given} that the trajectory eventually visits the future state $s_k$.

According to HCA theory ~\citep{hca}, we can construct an unbiased estimate of the Q-value. This estimate re-weights future returns using the importance ratio between the hindsight distribution and the policy distribution:
\begin{equation}
\label{eq:hca_general}
\begin{split}
Q(s_t, a) \approx & \ \hat{r}(s_t, a) + \sum_{k=t+1}^{T-1} \gamma^{k-t} \frac{h(a|s_t, s_k)}{\pi(a|s_t)} R_k \\
& + \gamma^{T-t} \frac{h(a|s_t, s_T)}{\pi(a|s_t)} V(s_T)
\end{split}
\end{equation}
where $\hat{r}$ is an estimate of the immediate reward, $R_k$ is the reward at step $k$, $\gamma$ is the discount factor  and $V$ is the state-value function.

While classical HCA requires training a separate parameterized model to estimate $h$ via supervised learning, in the context of LLM-based agents, we can leverage the inherent reasoning capabilities of the agent itself. Instead of explicit training, we simulate the hindsight distribution by injecting the realized outcome (posterior information) directly into the agent's input context via prompting. By conditioning the LLM on the future state, the model can effectively approximate the posterior probability $P(a_t|s_t, s_k)$, using its world knowledge to identify critical actions that causally lead to the observed outcome.
\section{HCAPO}
\label{sec:method}

\begin{figure*}[t]
    \centering
    \begin{subfigure}[b]{0.49\textwidth}
        \centering
        \includegraphics[width=\linewidth]{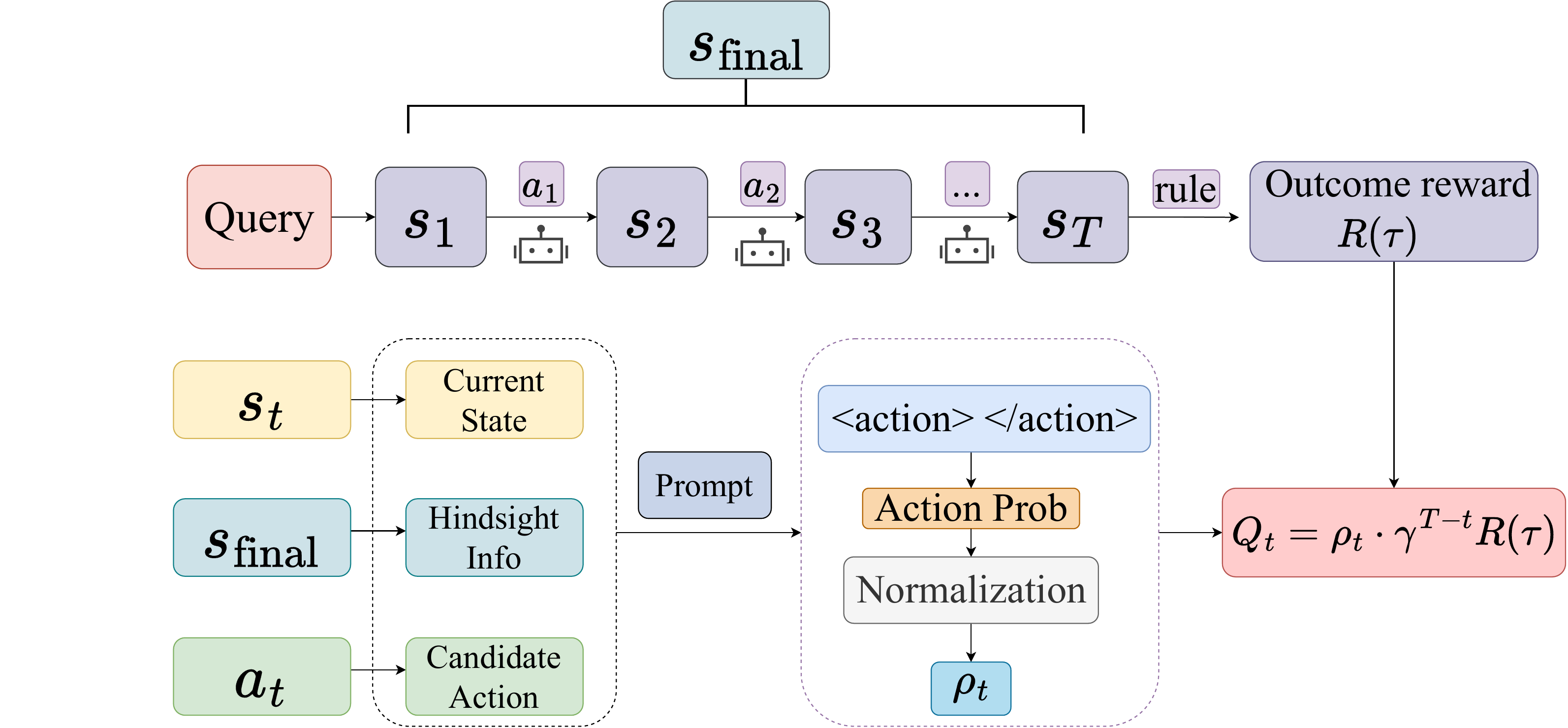} 
        \label{fig:step_logic}
    \end{subfigure}
    \hfill
    \begin{subfigure}[b]{0.5\textwidth}
        \centering
        \includegraphics[width=\linewidth]{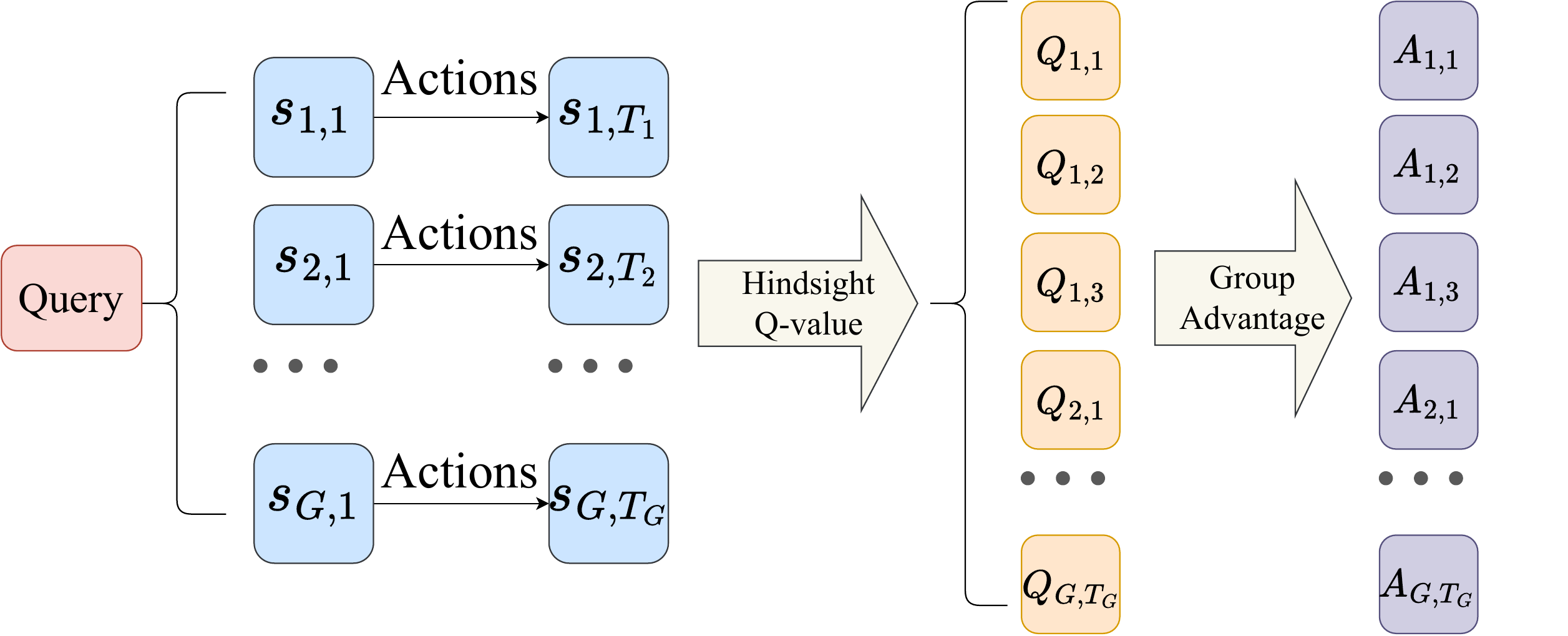} 
        \label{fig:group_flow}
    \end{subfigure}
    
    \caption{The HCAPO framework. (a) Illustrates the generative verification process: for a candidate action $a_t$, the LLM acts as a critic to compute the hindsight score $\rho_t$ by conditioning on the state $s_t$ and hindsight information $s_{final}$. (b) Shows the full optimization loop where a group of $G$ trajectories is evaluated via Hindsight Q-values to produce the final group-based advantage $A_{i,t}$.}
    \label{fig:overall_framework}
\end{figure*}

We introduce \textbf{Hindsight Credit Assignment Policy Optimization (HCAPO)}, a value-free reinforcement learning framework designed to resolve the sparse-reward bottleneck in long-horizon LLM agent tasks. HCAPO refines the coarse trajectory-level feedback into a fine-grained, step-level advantage signal by leveraging the agent's intrinsic reasoning capabilities. The overall framework is illustrated in Figure~\ref{fig:overall_framework}.
\subsection{Refined Hindsight Q-Value for Sparse Rewards}
\label{sec:q_derivation}

Standard value-free methods like GRPO \citep{deepseekmath} suffer from credit assignment challenges in long-horizon tasks, as they uniformly assign the terminal reward $R(\tau_i)$ to every action in the trajectory. This fails to distinguish pivotal state-action pairs from redundant steps. To resolve this, we derive a refined Hindsight Q-value grounded in HCA theory.

In tasks characterized by sparse terminal rewards (where intermediate rewards $r_{t<T}=0$), the HCA formulation \citep{hca} simplifies significantly. We define the refined Q-value for action $a_t$ at state $s_t$ as:
\begin{equation}
\label{eq:refined_q}
\begin{aligned}
    Q^{\text{H}}_{i,t} &= \rho_{i,t} \cdot G_{i,t}, \\  \rho_{i,t}& = \frac{h(a_t \mid s_t, s_{\text{final}})}{\pi(a_t \mid s_t)}
    \end{aligned}
\end{equation}
Here, $G_{i,t} = \gamma^{T-t} R(\tau_i)$ represents the discounted future return, and $\rho_{i,t}$ is the hindsight importance ratio. This ratio acts as a ``causal filter'': if the action's probability increases when conditioned on the successful outcome, its credit is amplified ($\rho_{i,t} > 1$); if it decreases, its credit is suppressed ($\rho_{i,t} < 1$). This mechanism effectively amplifies the credit for actions that are significantly more likely to occur given the knowledge of the successful outcome ($s_{\mathrm{final}}$).

\subsection{Generative Verification and Ratio Estimation}
\label{sec:bayesian_estimation}

Implementing the hindsight importance ratio $\rho = h/\pi$ in LLM agents faces two major obstacles. First, the prior policy $\pi(a_t|s_t)$ is intractable due to the vast, combinatorial nature of natural language action spaces. Second, classical HCA theory \citep{hca} requires training a separate parameterized model to approximate the hindsight distribution $h$. 

We resolve the two obstacles by leveraging the LLM's inherent reasoning capabilities through Generative Verification. Instead of training a new model, we``simulate" the hindsight distribution by injecting the successful outcome $s_{\text{final}}$ directly into the model's prompt. To estimate the ratio $\rho$ without explicit knowledge of the action space, we establish a link through a Bayesian lens. By the Law of Total Probability, the prior is the marginalization of the posterior over all potential outcomes: $\pi(a_t|s_t) = \mathbb{E}_{s_{\text{final}}} [\pi(a_t \mid s_t, s_{\text{final}})]$. 

Specifically, let action $a_t$ consist of tokens $(y_1, \dots, y_{|a_t|})$. We first compute $\pi_{\text{hind}}(a_t)$ as the exponential of the mean log-probabilities, conditioned on the successful state:
\begin{equation}
\label{eq:score_calc}
    \pi_{\text{hind}}(a_t) = \exp \left( \frac{1}{T_{\text{temp}} |a_t|} \sum_{j=1}^{|a_t|} \log \pi_\theta(y_j \mid y_{<j}, s_t, s_{\text{final}}) \right)
\end{equation}
where $T_{\text{temp}}$ is a sharpening temperature. By the Law of Total Probability, the prior policy is the marginalization of the posterior over all possible outcomes: $\pi(a_t|s_t) = \mathbb{E}_{s_{\text{final}}} [\pi(a_t \mid s_t, s_{\text{final}})]$. Since this marginalization is intractable, we approximate it using the empirical mean of hindsight scores within a trajectory, $\bar{\pi}_{\text{hind}}$, which serves as a robust surrogate for the prior. This leads to a self-normalized importance ratio estimator:
\begin{equation}
\label{eq:ratio_calc}
\begin{aligned}
    \rho_{t} =& \text{clip}\left( \frac{\pi_{\text{hind}}(a_{t})}{\bar{\pi}_{\text{hind}}}, C_{\min}, C_{\max} \right), \\
   & \bar{\pi}_{\text{hind}} = \frac{1}{T} \sum_{k=1}^{T} \pi_{\text{hind}}(a_k)
\end{aligned}
\end{equation}
This self-normalized approach transforms the intractable posterior estimation into a tractable scoring task. In long-horizon agent tasks such as ALFWorld, critical decision nodes may involve multiple consecutive actions; the intra-trajectory normalization over $\bar{\pi}_{\text{hind}}$ provides a meaningful local reference, akin to group-normalization across actions within the same episode. This enables efficient credit assignment without external models.
\subsection{Multi-Scale Optimization}
\label{sec:optimization_rationale}

HCAPO integrates two complementary scales of feedback: a \textbf{macro-scale} outcome signal for global stability and a \textbf{micro-scale} hindsight signal for local precision. The final composite advantage for the $i$-th trajectory in a group of size $G$ is:
\begin{equation}
\label{eq:hcapo_advantage}
A_{i,t}^{\text{HCAPO}} = \underbrace{\frac{R(\tau_i) - \mu_{R}}{\sigma_{R}}}_{\text{Macro (GRPO)}} + \omega \cdot \underbrace{\frac{Q^{\text{H}}_{i,t} - \mu_{\text{H}}}{\sigma_{\text{H}}}}_{\text{Micro (Hindsight)}}
\end{equation}
where $\mu_{\text{H}}$ and $\sigma_{\text{H}}$ are the group statistics of $Q^{\text{H}}$ at time step $t$. We argue that this cross-state normalization is theoretically sound for bottleneck learning in Section~\ref{sec:theoretical_analysis}.

Specifically, we apply a ``do-no-harm" protective mask to zero out negative hindsight signals in successful trials. The policy $\pi_\theta$ is optimized using the PPO \citep{ppo} surrogate objective (Eq. \ref{eq:final_objective}).

\begin{equation}
\label{eq:final_objective}
\begin{aligned}
    &\mathcal{J}(\theta) = \mathbb{E}_{\{\tau_i\}_{i=1}^K \sim \pi_{\theta_{\text{old}}}} \Bigg[ \frac{1}{K} \sum_{i=1}^K \frac{1}{T_i} \sum_{t=1}^{T_i} \min \Big( r_{i,t}(\theta) A_{i,t}^{\text{HCAPO}}, \\
    &\text{clip}(r_{i,t}(\theta), 1-\epsilon, 1+\epsilon) A_{i,t}^{\text{HCAPO}} \Big) - \beta_{\text{KL}} \mathbb{D}_{\text{KL}}(\pi_\theta || \pi_{\text{ref}}) \Bigg]
\end{aligned}
\end{equation}

where $\epsilon$ is the clipping parameter to constrain policy updates, and $\beta_{\text{KL}}$ penalizes the KL divergence against the reference policy $\pi_{\text{ref}}$ to prevent model collapse.

To further stabilize credit assignment in tasks with rigid causal chains, we optionally apply a \textbf{temporal smoothing} mechanism to $Q^{\text{H}}_{i,t}$ to distribute credit across adjacent reasoning and action steps (see Appendix~\ref{app:temporal_smoothing}).

The pseudocode for HCAPO is provided in Appendix~\ref{app:pseudocode}.
\section{Theoretical Rationale for HCAPO}
\label{sec:theoretical_analysis}

In this section, we provide a formal analysis for the synergy between macro-scale outcome signals and micro-scale hindsight guidance. We demonstrate that HCAPO's composite advantage effectively resolves the credit assignment problem by targeting task bottlenecks while maintaining overall training stability.

\subsection{Synergy of Macro and Micro Advantages}
The composite advantage in HCAPO integrates two complementary scales of feedback. Recalling Eq.~\ref{eq:hcapo_advantage}, the total advantage $A_{i,t}^{\text{HCAPO}}$ is formulated as:
\begin{equation}
    A_{i,t}^{\text{HCAPO}} = \underbrace{A_i^{\text{GRPO}}}_{\text{Macro Signal}} + \omega \cdot \underbrace{\frac{Q^{\text{H}}_{i,t} - \mu_{\text{H}}}{\sigma_{\text{H}}}}_{\text{Micro Correction}}
\end{equation}

\paragraph{Macro Stability from GRPO.} 
The macro signal, derived from standard trajectory-level GRPO, provides a robust and consistent reinforcement signal.It ensures that the policy trends toward high-reward outcomes. However, this signal assigns credit uniformly to all actions in a successful trial, regardless of their actual contribution.

\paragraph{Micro Precision from HCAPO.} 
The micro correction term acts as a high-resolution "filter" specifically designed for critical decision nodes. While the macro signal maintains the global task direction, the hindsight-refined $Q^{\text{H}}$ isolates the causal contribution of individual actions. This allows the model to amplify credit for pivotal "breakthrough" decisions while suppressing the influence of redundant or noisy steps that happened to occur on the path to success.

\subsection{Rationale for Cross-State Normalization}
A potential concern is the use of a global group mean $\mu_H$ computed across heterogeneous states. The global group mean $\mu_H$ converges to the expectation of $Q^{\text{H}}$ over the sampled state-action pairs across all trajectories in the group:
\begin{equation}
    \mu_{\text{H}}\approx \mathbb{E}_{s \sim d^\pi, a \sim \pi} [Q^{\text{H}}(s, a)].
\end{equation}
Applying the Law of Total Expectation, we can decompose this global expectation:
\begin{equation}
    \mu_{\text{H}} \approx \mathbb{E}_{s \sim d^\pi} \left[ \mathbb{E}_{a \sim \pi(a|s)} [Q^{\text{H}}(s, a) \mid s] \right] = \mathbb{E}_{s \sim d^\pi} [V^{\text{H}}(s)],
\end{equation}
where $V^{\text{H}}(s)$ is the hindsight state-value function and $d^\pi(s)$ is the state visitation distribution under the current policy. This proves that $\mu_{\text{H}}$ is a non-parametric estimate of the average expected utility across the entire task-visitation spaces.

The core strength of HCAPO lies in its ability to automatically identify task bottlenecks. Let $s^*$ be a pivotal bottleneck state. Before the breakthrough ($s < s^*$), the value is low ($V_{\text{low}}$); after the breakthrough ($s > s^*$), the value significantly increases ($V_{\text{high}}$). 

Since the global mean $\mu_{\text{H}}$ means the expectation over all states, it naturally falls between these two regimes: $V_{\text{low}} < \mu_{\text{H}} < V_{\text{high}}$. This positioning makes $\mu_{\text{H}}$ an ideal adaptive threshold for credit assignment at the bottleneck $s^*$:
\begin{itemize}
    \item \textbf{Breakthrough Actions ($a^*$):} Lead to $Q^{\text{H}} \approx V_{\text{high}}$, resulting in a large positive advantage ($V_{\text{high}} - \mu_{\text{H}} > 0$).
    \item \textbf{Non-instrumental Actions ($a^-$):} Result in $Q^{\text{H}} \approx V_{\text{low}}$, leading to a negative advantage ($V_{\text{low}} - \mu_{\text{H}} < 0$).
\end{itemize}
Similar to how GRPO works, HCAPO reduces variance for the bottleneck states. It filters out task-level background noise, enabling the agent to concentrate its learning capacity on triggering the transition from $V_{\text{low}}$ to $V_{\text{high}}$.

\paragraph{Summary.} In summary, HCAPO addresses the two fundamental limitations of standard group optimization: the coarse estimation of step-level Q-values and the misalignment of baselines for intermediate states. We resolve the former by refining Q-values through hindsight reasoning to isolate instrumental actions. For the latter, we demonstrate that HCAPO's multi-scale advantage integration provides a discriminative and accurate value estimate specifically at critical bottleneck nodes, while leveraging robust trajectory-level signals to maintain global training stability.
\section{Experiments}
\label{sec:experiment}

In this section, we present empirical evaluations of HCAPO across diverse agentic tasks. Specifically, we aim to demonstrate: (1) the superior capability of HCAPO in training LLM agents compared to trajectory-level baselines; (2) the behavioral evolution of agents regarding trajectory efficiency; and (3) the computational budget of our framework.

\subsection{Experiment Setup}

\paragraph{Benchmarks.} 
To ensure a rigorous comparison, our experimental setup and benchmarks follow all the configurations in GiGPO ~\citep{gigpo}. We first evaluate on \textbf{ALFWorld} ~\citep{alfworld}, an embodied environment assessing multi-step reasoning across six categories of household tasks. Secondly, we use \textbf{WebShop} ~\citep{webshop}, a web-based environment where agents navigate HTML sites to purchase items matching specific attributes; we report both the average Score and the Success Rate, which respectively capture the quality of task completion. Finally, we evaluate on \textbf{Search-augmented QA} tasks, including single-hop (NQ~\citep{nq}, TriviaQA~\citep{triviaqa}, PopQA~\citep{popqa}) and multi-hop (HotpotQA~\citep{hotpotqa}, 2Wiki~\citep{2wiki}, MuSiQue~\citep{trivedi2022musique}, Bamboogle~\citep{Bamboogle}) datasets. We treat NQ and HotpotQA as in-domain benchmarks and use the remaining datasets to assess out-of-domain generalization.

\paragraph{Baselines.} 
To ensure a standardized and rigorous comparison, we adopt the baseline results directly as reported in the original GiGPO paper ~\citep{gigpo} and EMPG paper ~\citep{empg}. For \textbf{ALFWorld} and \textbf{WebShop}, the baselines include: (1) \textit{Closed-source LLMs}: GPT-4o~\citep{gpt4} and Gemini-2.5-Pro~\citep{team2023gemini}; (2) \textit{Prompting agents}: ReAct~\citep{react} and Reflexion~\citep{reflexion}; and (3) \textit{RL training methods}: PPO~\citep{ppo}, RLOO~\citep{rloo}, GRPO ~\citep{deepseekmath},EMPG ~\citep{empg} and the state-of-the-art GiGPO ~\citep{gigpo}. 

For \textbf{search-augmented QA} tasks, following the experimental protocol in GiGPO ~\citep{gigpo}, we compare HCAPO against a specific suite of baselines including R1-Instruct, Search-R1~\citep{searchr1}, ZeroSearch~\citep{zerosearch}, StepSearch~\citep{zerosearch} and the state-of-the-art GiGPO~\citep{gigpo}. By utilizing the figures reported in the prior literature, we ensure that our evaluation is strictly consistent with the existing state-of-the-art benchmarks and maintains a fair comparison across all multi-turn reasoning and tool-calling tasks.

\paragraph{Training Details.} 
We utilize the Qwen2.5-Instruct series (1.5B, 3B, and 7B) \citep{qwen2.5} as our base models. To ensure a fair comparison, all experimental settings are kept identical to those in GiGPO \citep{gigpo}. Detailed settings are provided in Appendix~\ref{app:experiment_details}.

\subsection{Performance on ALFWorld and WebShop}


\begin{table*}[t]
\centering
\caption{Performance on ALFWorld and WebShop. Results are averaged over 3 random seeds. For ALFWorld, we report the average success rate (\%) for each subtask as well as the overall result. For WebShop, we report both the average score and the average success rate (\%). We compare our proposed \textbf{HCAPO} with GRPO and GiGPO. Best results are \textbf{bolded}.}
\label{tab:main_results}
\resizebox{\textwidth}{!}{
\begin{tabular}{ll cccccc c cc}
\toprule
\multirow{2}{*}{\textbf{Type}} & \multirow{2}{*}{\textbf{Method}} & \multicolumn{7}{c}{\textbf{ALFWorld}} & \multicolumn{2}{c}{\textbf{WebShop}} \\
\cmidrule(lr){3-9} \cmidrule(lr){10-11}
& & \textbf{Pick} & \textbf{Look} & \textbf{Clean} & \textbf{Heat} & \textbf{Cool} & \textbf{Pick2} & \textbf{All} & \textbf{Score} & \textbf{Succ.} \\
\midrule
\multicolumn{11}{l}{\textit{Closed-Source Model}} \\
\midrule
Prompting & GPT-4o & 75.3 & 60.8 & 31.2 & 56.7 & 21.6 & 49.8 & 48.0 & 31.8 & 23.7 \\
Prompting & Gemini-2.5-Pro & 92.8 & 63.3 & 62.1 & 69.0 & 26.6 & 58.7 & 60.3 & 42.5 & 35.9 \\
\midrule
\multicolumn{11}{l}{\textit{Qwen2.5-1.5B-Instruct}} \\
\midrule
Prompting & Qwen2.5 & 5.9 & 5.5 & 3.3 & 9.7 & 4.2 & 0.0 & 4.1 & 23.1 & 5.2 \\
Prompting & ReAct & 17.4 & 20.5 & 15.7 & 6.2 & 7.7 & 2.0 & 12.8 & 40.1 & 11.3 \\
Prompting & Reflexion & 35.3 & 22.2 & 21.7 & 13.6 & 19.4 & 3.7 & 21.8 & 55.8 & 21.9 \\
RL Training & PPO (with critic) & 64.8$\pm$3.5 & 40.5$\pm$6.9 & 57.1$\pm$4.9 & 60.6$\pm$6.6 & 46.4$\pm$4.0 & 47.4$\pm$1.9 & 54.4$\pm$3.1 & 73.8$\pm$3.0 & 51.5$\pm$2.9 \\
RL Training & RLOO & 88.3$\pm$3.0 & 52.8$\pm$8.6 & 71.0$\pm$5.9 & 62.8$\pm$8.7 & 66.4$\pm$5.5 & 56.9$\pm$4.7 & 69.7$\pm$2.5 & 73.9$\pm$5.6 & 52.1$\pm$6.7 \\
\cmidrule(lr){2-11}
RL Training & GRPO & 85.3$\pm$1.5 & 53.7$\pm$8.0 & 84.5$\pm$6.8 & 78.2$\pm$7.9 & 59.7$\pm$5.0 & 53.5$\pm$5.6 & 72.8$\pm$3.6 & 75.8$\pm$3.5 & 56.8$\pm$3.8 \\
RL Training & EMPG & 85.5 & 33.5 & 78.9 & 76.2 & 74.7 & 69.1 & 73.7 & 80.4 & 60.8 \\
RL Training & GiGPO & \textbf{94.4$\pm$5.9} & 67.5$\pm$4.6 & 94.8$\pm$3.8 & \textbf{94.4$\pm$7.8} & 79.8$\pm$4.7 & \textbf{76.4$\pm$5.4} & 86.7$\pm$1.7 & 83.1$\pm$1.6 & 65.0$\pm$3.2 \\
RL Training & \textbf{HCAPO (Ours)} & 88.6$\pm$7.0 & \textbf{75.0$\pm$0.0} & \textbf{97.6$\pm$1.8} & 90.7$\pm$6.9 & \textbf{84.2$\pm$0.0} & 74.2$\pm$6.9 & \textbf{87.0$\pm$4.1} & \textbf{83.8$\pm$0.7} & \textbf{68.5$\pm$1.0} \\
\midrule
\multicolumn{11}{l}{\textit{Base Model: Qwen2.5-7B-Instruct}} \\
\midrule
Prompting & Qwen2.5 & 33.4 & 21.6 & 19.3 & 6.9 & 2.8 & 3.2 & 14.8 & 26.4 & 7.8 \\
Prompting & ReAct & 48.5 & 35.4 & 34.3 & 13.2 & 18.2 & 17.6 & 31.2 & 46.2 & 19.5 \\
Prompting & Reflexion & 62.0 & 41.6 & 44.9 & 30.9 & 36.3 & 23.8 & 42.7 & 58.1 & 28.8 \\
RL Training & PPO (with critic) & 92.3$\pm$4.0 & 64.0$\pm$8.4 & 92.5$\pm$2.4 & \textbf{89.5}$\pm$\textbf{7.0} & 80.3$\pm$2.0 & 68.8$\pm$8.3 & 80.4$\pm$2.7 & 81.4$\pm$3.1 & 68.7$\pm$5.1 \\
RL Training & RLOO & 87.6$\pm$4.3 & 78.2$\pm$8.3 & 87.3$\pm$5.8 & 81.3$\pm$7.6 & 71.9$\pm$5.2 & 48.9$\pm$8.4 & 75.5$\pm$4.6 & 80.3$\pm$3.2 & 65.7$\pm$4.0 \\
\cmidrule(lr){2-11}
RL Training & GRPO & 90.8$\pm$5.1 & 66.1$\pm$6.7 & 89.3$\pm$5.4 & 74.7$\pm$6.9 & 72.5$\pm$5.4 & 64.7$\pm$7.3 & 77.6$\pm$5.2 & 79.3$\pm$2.8 & 66.1$\pm$3.7 \\
RL Training & EMPG & 92.9 & 75.2 & 74.8 & 86.3 & 73.7 & 65.3 & 78.5 & 81.0 & 69.3 \\
RL Training & GiGPO & 97.7$\pm$1.6 & 82.7$\pm$7.9 & \textbf{98.8$\pm$1.6} & 83.7$\pm$7.2 & 89.3$\pm$8.2 & 79.2$\pm$6.6 & 90.8$\pm$1.3 & 84.4$\pm$2.9 & 72.8$\pm$3.2 \\
RL Training & \textbf{HCAPO (Ours)} & \textbf{99.1$\pm$1.3} & \textbf{90.3$\pm$2.0} & 97.3$\pm$1.9 & 81.8$\pm$8.8 & \textbf{90.8$\pm$6.6} & \textbf{81.9$\pm$10.0} & \textbf{91.4$\pm$2.3} & \textbf{85.1$\pm$1.3} & \textbf{73.8$\pm$2.8} \\
\bottomrule
\end{tabular}
}
\end{table*}

As shown in Table~\ref{tab:main_results}, HCAPO achieves significant gains over the trajectory-level baseline GRPO and demonstrates performance comparable to the state-of-the-art GiGPO across both ALFWorld and WebShop. On ALFWorld (7B), HCAPO reaches an overall success rate of \textbf{91.4\%}, surpassing GRPO's 77.6\% by 13.8 points and slightly exceeding GiGPO (90.8\%). Similar gains are observed at 1.5B (87.0\% vs. 72.8\% for GRPO). On WebShop, HCAPO improves both evaluation metrics: at 7B, the average Score rises from 79.3 to 85.1 and the Success Rate from 66.1 to 73.8; at 1.5B, Score increases from 75.8 to 83.8 and Success Rate from 56.8 to 68.5, closely matching or exceeding GiGPO.

These results highlight that HCAPO effectively overcomes the limitations of coarse trajectory-level feedback. While GRPO struggles to isolate instrumental actions in long interaction sequences, HCAPO's hindsight ratio successfully identifies key actions even in complex environments like \textit{Pick2} or \textit{Cool}, leading to more robust and effective learning.

Furthermore, HCAPO's performance becomes more stable as the model scales from 1.5B to 7B. This scaling trend suggests that larger models are better equipped to leverage hindsight information, likely due to their superior reasoning capacity and instruction-following abilities, which allow for more consistent posterior evaluations of past actions.Additional ablation results are reported in Appendix~\ref{app:ablation_omega}.

\subsection{Performance on Search-augmented QA Tasks}

\begin{table*}[t]
\centering
\caption{Performance on search-augmented QA tasks. $\dagger$ and $\star$ indicate in-domain and out-of-domain datasets, respectively. Bold indicates the best performance in each category.}
\label{tab:qa_results}
\resizebox{\textwidth}{!}{
\begin{tabular}{ll ccc cccc c}
\toprule
\multirow{2}{*}{\textbf{Type}} & \multirow{2}{*}{\textbf{Method}} & \multicolumn{3}{c}{\textbf{Single-Hop QA}} & \multicolumn{4}{c}{\textbf{Multi-Hop QA}} & \multirow{2}{*}{\textbf{Avg.}} \\
\cmidrule(lr){3-5} \cmidrule(lr){6-9}
& & \textbf{NQ}$\dagger$ & \textbf{TriviaQA}$\star$ & \textbf{PopQA}$\star$ & \textbf{HotpotQA}$\dagger$ & \textbf{2Wiki}$\star$ & \textbf{MuSiQue}$\star$ & \textbf{Bamboogle}$\star$ & \\
\midrule
\multicolumn{10}{l}{\textit{Base Model: Qwen2.5-3B-Instruct}} \\
\midrule
RL Training & R1-Instruct & 27.0 & 53.7 & 19.9 & 23.7 & 29.2 & 7.2 & 29.3 & 27.1 \\
RL Training & Search-R1 & 34.1 & 54.5 & 37.8 & 32.4 & 31.9 & 10.3 & 26.4 & 32.5 \\
RL Training & ZeroSearch & 41.4 & 57.4 & \underline{44.8} & 27.4 & 30.0 & 9.8 & 11.1 & 31.7 \\
RL Training & StepSearch & -- & -- & -- & 34.5 & 32.0 & \textbf{17.4} & -- & 34.4 \\
RL Training & GiGPO & \underline{42.0} & \underline{59.5} & 42.4 & \underline{36.9} & \textbf{37.0} & 12.6 & \underline{64.1} & \underline{42.1} \\
\cmidrule(lr){2-10}
RL Training & \textbf{HCAPO (Ours)} & \textbf{44.4} & \textbf{60.5} & \textbf{45.5} & \textbf{38.6} & \underline{36.3} & \underline{14.8} & \textbf{64.5} & \textbf{44.3} \\
\midrule
\multicolumn{10}{l}{\textit{Base Model: Qwen2.5-7B-Instruct}} \\
\midrule
RL Training & R1-Instruct & 21.0 & 44.9 & 17.1 & 20.8 & 27.5 & 6.0 & 19.2 & 22.4 \\
RL Training & Search-R1 & 39.3 & 61.0 & 39.7 & 37.0 & 40.1 & 14.6 & 36.8 & 38.5 \\
RL Training & ZeroSearch & 43.6 & 61.8 & \textbf{51.5} & 34.6 & 35.2 & 18.4 & 27.8 & 39.1 \\
RL Training & StepSearch & -- & -- & -- & 38.6 & 36.6 & \textbf{22.6} & -- & 40.0 \\
RL Training & GiGPO & \textbf{46.4} & \underline{64.7} & 46.1 & \underline{41.6} & \textbf{43.6} & \underline{18.9} & \underline{68.9} & \underline{47.2} \\
\cmidrule(lr){2-10}
RL Training & \textbf{HCAPO (Ours)} & \underline{46.1} & \textbf{65.5} & \underline{47.6} & \textbf{42.1} & \underline{43.1} & 17.7 & \textbf{69.0} & \textbf{48.3} \\
\bottomrule
\end{tabular}
}
\end{table*}

Table~\ref{tab:qa_results} presents the results on search-augmented QA tasks. We observe that HCAPO achieves strong and consistent gains across both single-hop and multi-hop reasoning datasets. Notably, HCAPO reaches an average success rate of \textbf{48.3\%} at 7B, outperforming prior strong baselines such as Search-R1 and StepSearch, and maintaining a performance level comparable to GiGPO. 

In Single-Hop QA, HCAPO yields consistent performance gains across the datasets, primarily because it more effectively identifies the specific query that provides the most critical information for the final answer. HCAPO successfully highlights the high-utility ``golden query" that leads directly to the correct evidence. By concentrating credit on these pivotal actions rather than distributing it uniformly across the interaction history, HCAPO reinforces the most efficient retrieval paths and enhances the agent's ability to locate core evidence in a single step.

\subsection{Dynamics of Behavioral Conciseness}
\label{sec:conciseness}

We investigate how the hindsight signal reshapes the agent's decision-making process during training. A unique advantage of HCAPO is its ability to identify and suppress redundant actions even within successful trajectories. We define ``redundant actions" as those assigned a hindsight confidence score $\pi_{\text{hind}} \le 0.9$ when $T_{temp}=1$, indicating low instrumental utility relative to the successful outcome.

\begin{figure}[t]
    \centering
    \includegraphics[width=\linewidth]{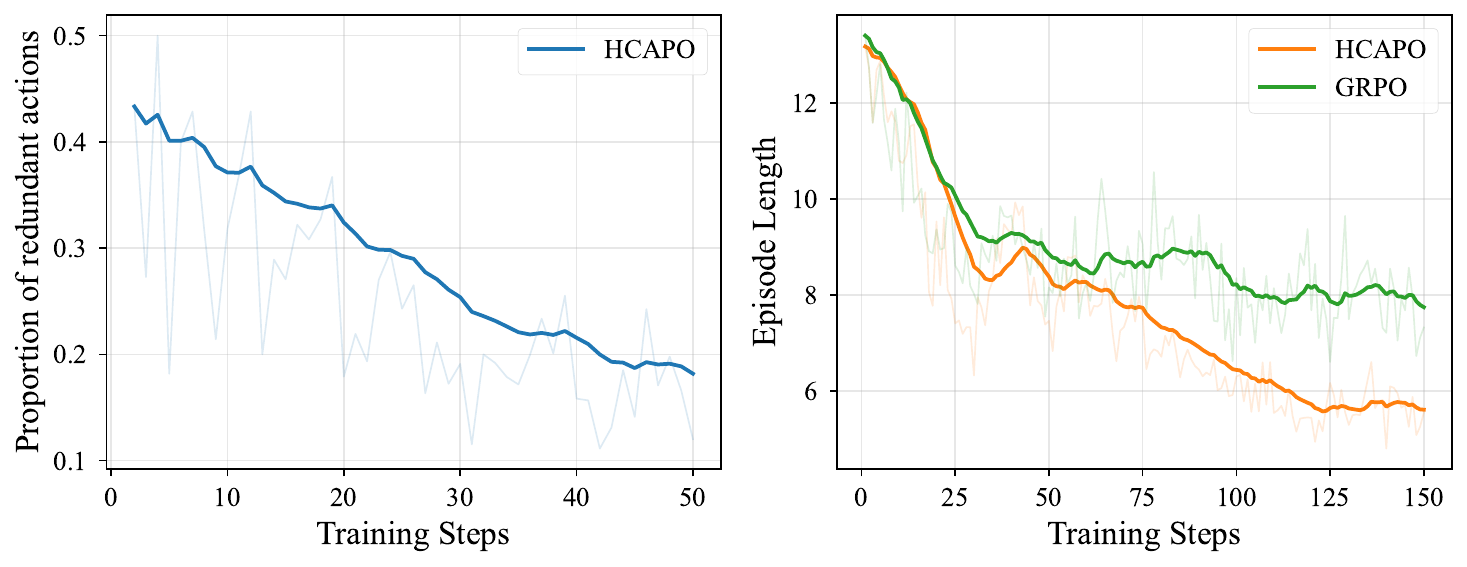}
    \caption{\textbf{LEFT:} Proportion of redundant actions during training in webshop task. \textbf{RIGHT:} Path-shortening effect of HCAPO vs. GRPO in webshop task.}
    \label{fig:evolution}
    \vspace{-5mm}
\end{figure}

As illustrated in Figure~\ref{fig:evolution}(a), we track the proportion of such redundant actions over the training process. Initially, the policy generates a high percentage of noise actions. However, as HCAPO penalizes these steps, the ``pruning rate" significantly improves, and the frequency of redundant actions steadily decreases. This signifies that the agent is successfully internalizing the essential causal logic of the task.Furthermore, these results provide strong evidence of HCAPO's discriminative power for noisy actions.

This behavioral refinement is further evidenced by the path-shortening effect shown in Figure~\ref{fig:evolution}(b). While the GRPO baseline maintains a high average trajectory length ($\approx 7.8$ steps) due to its inability to distinguish key actions, HCAPO agents converge to a more concise policy ($\approx 5.8$ steps).

\subsection{Analysis of Computational Efficiency}
\label{sec:efficiency}

In this section, we analyze the computational overhead introduced by the hindsight mechanism. The primary addition to the training pipeline is the Generative Verification process used to compute the hindsight probability $\pi_{\text{hind}}$.

\paragraph{Efficiency of Generative Verification.} 
Crucially, Generative Verification is computationally efficient by design. Unlike the Generation phase, which requires time-consuming auto-regressive decoding to generate actions token-by-token, Generative Verification only involves scoring existing trajectories. The model extracts the log-probabilities of the action tokens in a single forward pass. This parallelizable, prefix-based computation bypasses the sequential bottleneck of auto-regressive decoding, making the Generative Verification phase significantly faster than the Generation phase.
\begin{figure}[t]
    \centering
    \includegraphics[width=1\linewidth]{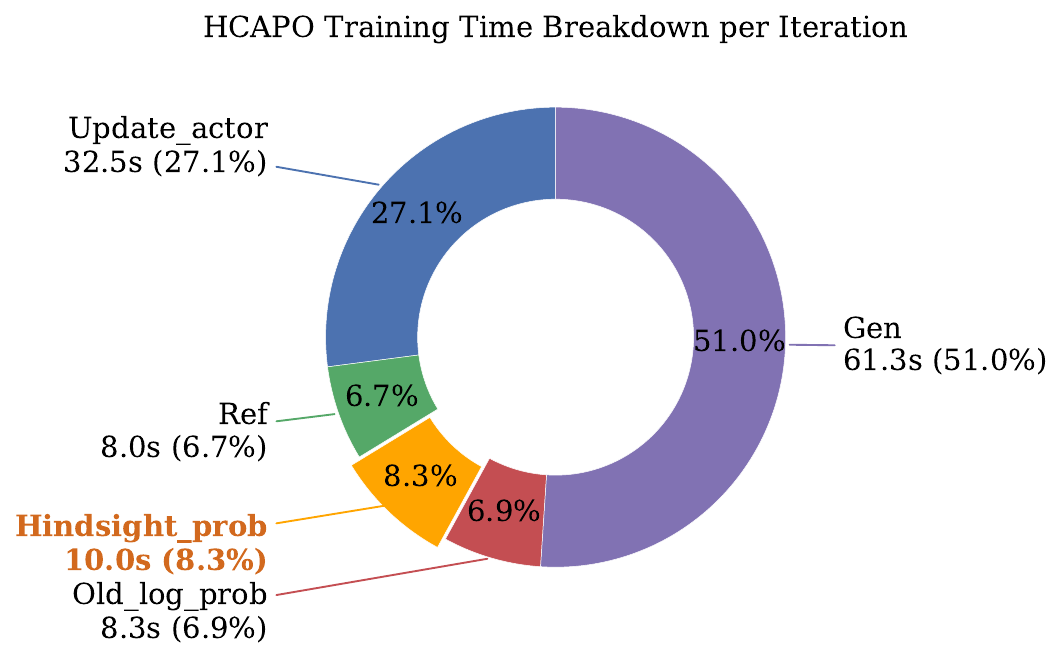}
    \caption{Computational cost breakdown during training. The hindsight audit pass accounts for only 8.3\% of total training time.}
    \label{fig:latency_pie}
\end{figure}
\paragraph{Latency Breakdown.}
Figure~\ref{fig:latency_pie} provides a breakdown of the step-wise training latency. As shown in the pie chart, the vast majority of computational resources are consumed by the Generation stages. Attributable to its non-generative nature, the computation of hindsight prob accounts for only approximately $8.3\%$ of the total training time. This result demonstrates that HCAPO provides a high performance-to-cost ratio, delivering substantial improvements in credit assignment with minimal additional computational burden.


\section{Conclusion and Limitations}
\label{sec:conclusion}

\paragraph{Conclusion.} We introduce \textbf{HCAPO}, a value-free framework that bridges HCA theory and long-horizon LLM agent optimization. Our analysis reveals that accurate estimation of step-level action values is important and enough for credit assignment, even when coupled with a simplified global group normalization. By leveraging the LLM's intrinsic reasoning for generative verification, HCAPO provides a novel and efficient approach for scalable agent optimization without relying on external models.

\paragraph{Limitations.}Despite its effectiveness, HCAPO relies on the base model's reasoning capacity, which may limit the precision of credit signals in small models. 
Furthermore, while striving to preserve the agent's decision-making process, the inclusion of hindsight information inevitably introduces some degree of out-of-distribution data. Future work could explore specialized fine-tuning to better align this hindsight reasoning with the policy.

\bibliography{example_paper}
\bibliographystyle{icml2026}

\newpage
\appendix
\onecolumn

\section{Temporal Smoothing For HCAPO}
\label{app:temporal_smoothing}

\subsection{Methodology}
In multi-step tasks like ALFWorld, we observe a "credit disconnection" problem: the LLM verifier easily identifies the final "CleanObject" action as useful, but sometimes assigns lower scores to early-stage "navigational" or "preparatory" actions (e.g., "GoToPlace", "OpenObject"). However, the final success is strictly contingent on these predecessors. 

By applying the temporal smoothing window $\tilde{Q}^{\text{H}}_{i,t} = \alpha Q^{\text{H}}_{i,t} + (1-\alpha) Q^{\text{H}}_{i,t+1}$, we effectively allow the "breakthrough signal" from the terminal step to flow backward. With $\alpha=0.5$, we treat the reasoning step and its immediate execution as a coherent functional unit. This prevents the policy from over-optimizing for the final reward while neglecting the prerequisite steps.
\subsection{Experiments}
\begin{figure}[ht]
    \centering
    \includegraphics[width=0.8\linewidth]{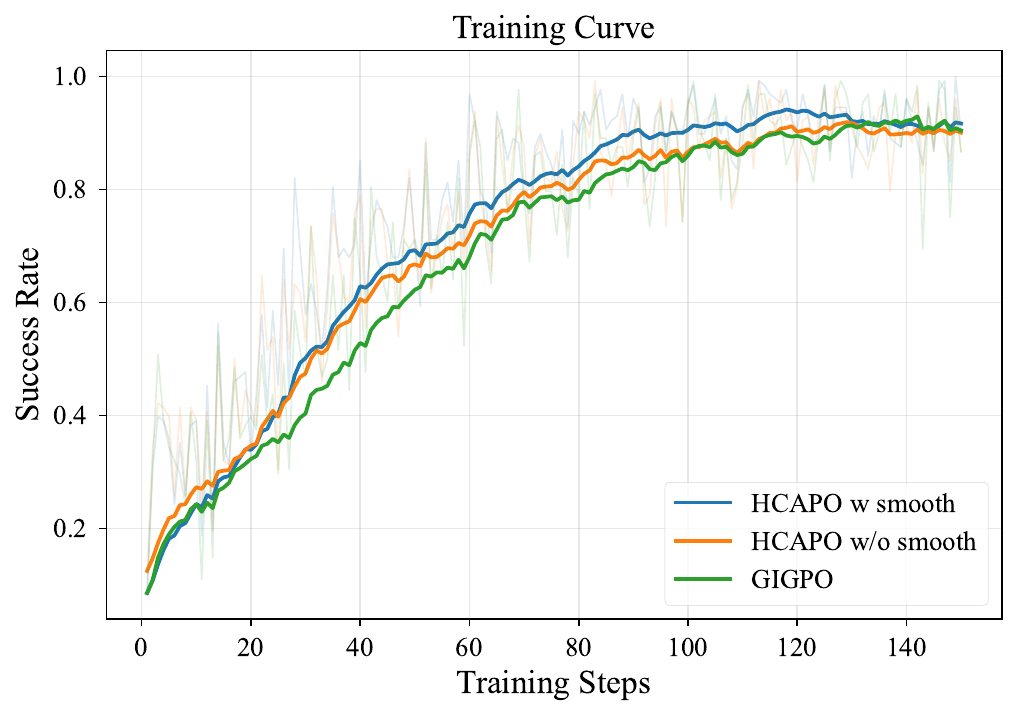}
    \caption{Success Rate during training}
    \label{fig:temporal_smoothing}
\end{figure}

Table~\ref{tab:temporal_smoothing} shows that smoothing significantly stabilizes learning in complex multi-step sequences, leading to higher overall success rates.Figure~\ref{fig:temporal_smoothing} illustrates the training stability improvement 
achieved by temporal smoothing in ALFWorld tasks.

\begin{table*}[ht]
\centering
\caption{Performance on ALFWorld with temporal smoothing. Results are averaged over 3 random seeds. For ALFWorld, we report the average success rate (\%) for each subtask as well as the overall result. We compare our proposed \textbf{HCAPO} with GRPO and GiGPO.}
\label{tab:temporal_smoothing}
\resizebox{\textwidth}{!}{
\begin{tabular}{ll cccccc c cc}
\toprule
\multirow{2}{*}{\textbf{Type}} & \multirow{2}{*}{\textbf{Method}} & \multicolumn{7}{c}{\textbf{ALFWorld}} \\
\cmidrule(lr){3-9} \cmidrule(lr){10-11}
& & \textbf{Pick} & \textbf{Look} & \textbf{Clean} & \textbf{Heat} & \textbf{Cool} & \textbf{Pick2} & \textbf{All} \\
\midrule
\multicolumn{11}{l}{\textit{Qwen2.5-1.5B-Instruct}} \\
\midrule
RL Training & GRPO & 85.3$\pm$1.5 & 53.7$\pm$8.0 & 84.5$\pm$6.8 & 78.2$\pm$7.9 & 59.7$\pm$5.0 & 53.5$\pm$5.6 & 72.8$\pm$3.6 \\
RL Training & GiGPO & \textbf{94.4$\pm$5.9} & 67.5$\pm$4.6 & 94.8$\pm$3.8 & \textbf{94.4$\pm$7.8} & 79.8$\pm$4.7 & 76.4$\pm$5.4 & 86.7$\pm$1.7 \\
RL Training & HCAPO w/o Smooth & 88.6$\pm$7.0 & \textbf{75.0$\pm$0.0} & \textbf{97.6$\pm$1.8} & 90.7$\pm$6.9 & 84.2$\pm$0.0 & 74.2$\pm$6.9 & 87.0$\pm$4.1 \\
RL Training & HCAPO w Smooth & 93.3$\pm$5.4 & 70.8$\pm$5.9 & 93.1$\pm$2.0 & 90.7$\pm$6.9 & \textbf{86.0$\pm$2.5} & \textbf{76.4$\pm$1.0} & \textbf{87.2$\pm$3.7} \\
\midrule
\multicolumn{11}{l}{\textit{Qwen2.5-7B-Instruct}} \\
\midrule
RL Training & GRPO & 90.8$\pm$5.1 & 66.1$\pm$6.7 & 89.3$\pm$5.4 & 74.7$\pm$6.9 & 72.5$\pm$5.4 & 64.7$\pm$7.3 & 77.6$\pm$5.2 \\
RL Training & GiGPO & 97.7$\pm$1.6 & 82.7$\pm$7.9 & \textbf{98.8$\pm$1.6} & 83.7$\pm$7.2 & 89.3$\pm$8.2 & 79.2$\pm$6.6 & 90.8$\pm$1.3 \\
RL Training & HCAPO w/o Smooth & 99.1$\pm$1.3 & 90.3$\pm$2.0 & 97.3$\pm$1.9 & 81.8$\pm$8.8 & 90.8$\pm$6.6 & 81.9$\pm$10.0 & 91.4$\pm$2.3 \\
RL Training & HCAPO w Smooth & \textbf{100.0$\pm$0.0} & \textbf{91.7$\pm$5.9} & 97.3$\pm$1.9 & \textbf{91.7$\pm$2.9} & \textbf{98.2$\pm$2.5} & \textbf{95.8$\pm$0.0} & \textbf{96.9$\pm$1.3} \\
\bottomrule
\end{tabular}
}
\end{table*}

\section{Algorithm Pseudocode}
\label{app:pseudocode}

\begin{algorithm}[H]
\caption{Training LLM Agents with HCAPO}
\label{alg:hcapo}
\begin{algorithmic}[1]
\STATE \textbf{Require:} Initial policy $\pi_\theta$, task distribution $p(X)$, weighting coefficient $\omega$, batch size $N$, clipping bounds $[C_{\min}, C_{\max}]$.
\FOR{each training iteration}
    \STATE Update old policy: $\theta_{\text{old}} \leftarrow \theta$
    
    \STATE \textit{// 1. Multi-step Rollout Phase}
    \STATE Sample task $x \sim p(X)$ and initialize $N$ identical environments.
    \FOR{$t = 1$ \text{to} $T$}
        \STATE Sample actions $a_{i,t} \sim \pi_{\theta_{\text{old}}}(\cdot \mid s_{i,t})$ for all $i \in \{1, \dots, N\}$.
        \STATE Execute actions, observation $\{o_{i,t}\}_{i=1}^N$ and then next states $\{s_{i,t+1}\}_{i=1}^N$.
    \ENDFOR
    
    \STATE \textit{// 2. Hindsight Credit Assignment Phase}
    \STATE Compute \textbf{Macro Advantage} $A_{i,t}^{\text{GRPO}}$ via trajectory-level relative rewards.
    \STATE Compute hindsight probabilities $\pi_{\text{hind}}(a_{i,t})$ via \textbf{Generative Verification}.
    \STATE Estimate importance ratios $\rho_{i,t} = \text{clip}(\pi_{\text{hind}}(a_{i,t}) / \bar{\pi}_{\text{hind}}, C_{\min}, C_{\max})$.
    \STATE Derive refined Hindsight Q-values $Q^{\text{H}}_{i,t} = \rho_{i,t} \cdot \gamma^{T-t} R(\tau_i)$.
    \STATE (Optional) Apply temporal smoothing: $\tilde{Q}^{\text{H}}_{i,t} = \alpha Q^{\text{H}}_{i,t} + (1-\alpha) Q^{\text{H}}_{i,t+1}$.
    \STATE Compute \textbf{Micro Advantage} via cross-state normalization: $A_{i,t}^{\text{Micro}} = \frac{Q^{\text{H}}_{i,t} - \mu_{\text{H}}}{\sigma_{\text{H}}}$.
    
    \STATE \textit{// 3. Policy Update Phase}
    \STATE Combine multi-scale advantages: $A_{i,t}^{\text{HCAPO}} = A_i^{\text{GRPO}} + \omega A_{i,t}^{\text{Micro}}$.
    \STATE Update policy $\theta$ by maximizing the PPO-clipped surrogate objective $\mathcal{J}_{\text{HCAPO}}(\theta)$.
\ENDFOR
\end{algorithmic}
\end{algorithm}

\section{Experiment Details}
\label{app:experiment_details}

\subsection{Details of Training}
\label{app:training_details}

\paragraph{Hyperparameters for ALFWorld.} 
All methods are configured with identical hyperparameters to ensure a fair comparison: the maximum prompt length is 2048 tokens, and the maximum response length is 512 tokens. Each episode allows up to 50 environment steps. The learning rate is set to $1\times 10^{-6}$ for the actor and $1\times 10^{-5}$ for the critic (used only in PPO baselines). We adopt a rule-based reward, assigning a reward of $+10$ for success and $0$ for failure. To handle invalid actions generated by the agent, we apply a reward penalty of $-0.1$. For all group-based RL methods (GRPO, GiGPO, HCAPO), we use a group size of $G=8$ and sample 16 different groups per rollout, resulting in a total of $16 \times 8 = 128$ environments. In contrast, PPO uses 128 separate environments for rollouts. The rollout temperature is set to 1.0, while the validation temperature is set to 0.4. The mini-batch size is 256, and the KL-divergence loss coefficient $\beta_{\text{KL}}$ is 0.01.

\paragraph{Hyperparameters for WebShop.} 
All methods are configured with the following hyperparameters: the maximum prompt length is 4096 tokens, and the maximum response length is 512 tokens. Each episode is limited to 15 environment steps. The learning rate is $1\times 10^{-6}$ for the actor and $1\times 10^{-5}$ for the critic. We adopt a rule-based reward, assigning a reward of $+10$ for success and $0$ for failure. Invalid actions are penalized with a reward of $-0.1$. As with ALFWorld, all group-based RL methods use a group size of $G=8$ and sample 16 groups per rollout, totaling 128 environments. PPO uses 128 distinct environments for rollouts. The rollout temperature is set to 1.0, and the validation temperature is 0.4. The mini-batch size is 64, and $\beta_{\text{KL}}$ is 0.01. 

\paragraph{Hyperparameters for Search-Augmented QA.} 
The maximum prompt length is 4096 tokens, and the maximum response length is 512 tokens. The maximum number of turns is set to 4. The learning rate is $1\times 10^{-6}$ for the actor. We adopt a rule-based reward, assigning a reward of $+1$ for success and $0$ for failure. Invalid actions are penalized with a reward of $-0.01$. We set the training data size to 256 and use a group size of $G=5$. Rollout and validation temperatures are set to 1.0 and 0.0, respectively. The mini-batch size is 512, and $\beta_{\text{KL}}$ is 0.001.

\paragraph{Computing Details.} 
For ALFWorld and WebShop, Qwen2.5-1.5B experiments are run on $4\times$ H20 GPUs and Qwen2.5-7B on $8\times$ H20 GPUs, each for 150 iterations. For search-augmented QA, Qwen2.5-3B uses $8\times$ H20 GPUs and Qwen2.5-7B uses $8\times$ H20 GPUs, each for 200 iterations.

\paragraph{HCAPO Specific Hyperparameters.} 
For the \textit{Generative Verification} process, we set the sharpening temperature $T_{\text{temp}} = 5.0$. The hindsight importance ratio $\rho_{i,t}$ is clipped within $[C_{\min}, C_{\max}] = [0.8, 1.2]$ to prevent training instability from extreme posterior estimations. The hindsight weighting coefficient $\omega$ is set to $1.0$. The temporal smoothing factor $\alpha$ is 0.5, and the discount factor $\gamma$ is 0.95.These hyperparameters are kept consistent across all benchmarks to demonstrate the robustness of the framework without the need for task-specific tuning.

\subsection{Agent Training Prompts}
\label{app:agent_prompts}

The prompts we use for LLM agents are constructed using Python-style string formatting, where placeholders enclosed in curly braces (\{\}) represent semantic slots. These placeholders, such as \texttt{\{task\_description\}}, \texttt{\{step\_count\}}, and \texttt{\{current\_observation\}}, are dynamically populated at runtime via Python’s \texttt{.format()} function. To enrich the agent’s context, we use historical information and set the history length to 2 for ALFWorld and WebShop and the full history for search-augmented QA experiments.

The \texttt{<think> </think>} block instructs the agent to explicitly perform step-by-step reasoning, thereby promoting Chain-of-Thought (CoT) style deliberation. The \texttt{<action> </action>} block is used to clearly indicate the final action decision. The search agent outputs reasoning traces within \texttt{<think> </think>}, issues search queries within \texttt{<search> </search>}, and provides final answers within \texttt{<answer> </answer>} tags. Retrieved evidence from the search engine is presented within \texttt{<information> </information>} tags. The detailed templates for each environment are provided below.

\subsubsection{ALFWorld Agent Training Template}
\begin{tcolorbox}[
    title=ALFWorld Agent Training Template,
    colback=gray!3, colframe=gray!75, sharp corners, boxrule=0.5pt, breakable
]
\begin{Verbatim}[breaklines=true, fontfamily=tt, fontsize=\small]
You are an expert agent operating in the ALFRED embodied Environment. 
Your task is to: {task_description}. 

Prior to this step, you have already taken {step_count} step(s). Below are the most recent {history_length} observations and the corresponding actions you took: 
{action_history}

You are now at step {current_step} and your current observation is: {current_observation}.
Your admissible actions of the current situation are: [{admissible_actions}].

Now it’s your turn to take an action. You should first reason step-by-step about the current situation. This reasoning process MUST be enclosed within <think> </think> tags. Once you’ve finished your reasoning, you should choose an admissible action for current step and present it within <action> </action> tags.
\end{Verbatim}
\end{tcolorbox}

\subsubsection{WebShop Agent Training Template}
\begin{tcolorbox}[
    title=WebShop Agent Training Template,
    colback=green!3, colframe=green!50!black, sharp corners, boxrule=0.5pt, breakable
]
\begin{Verbatim}[breaklines=true, fontfamily=tt, fontsize=\small]
You are an expert autonomous agent operating in the WebShop e-commerce environment. 
Your task is to: {task_description}. 

Prior to this step, you have already taken {step_count} step(s). Below are the most recent {history_length} observations and the corresponding actions you took: 
{action_history}

You are now at step {current_step} and your current observation is: {current_observation}.
Your admissible actions for the current situation are: [{available_actions}].

Now it’s your turn to take one action for the current step. You should first reason step-by-step about the current situation, then think carefully which admissible action best advances the shopping goal. This reasoning process MUST be enclosed within <think> </think> tags. Once you’ve finished your reasoning, you should choose an admissible action for current step and present it within <action> </action> tags.
\end{Verbatim}
\end{tcolorbox}

\subsubsection{Search-augmented QA Agent Training Template}
\begin{tcolorbox}[
    title=Search-augmented QA Agent Training Template,
    colback=blue!3, colframe=blue!75!black, sharp corners, boxrule=0.5pt, breakable
]
\begin{Verbatim}[breaklines=true, fontfamily=tt, fontsize=\small]
You are an expert agent tasked with answering the given question step-by-step. 
Your question: {task_description}. 

Prior to this step, you have already taken {step_count} step(s). Below is the interaction history where <search> </search> wrapped your past search queries and <information> </information> wrapped the corresponding search results returned by the external search engine. 
History: {memory_context}

Now it’s your turn to respond for the current step. You should first conduct reasoning process. This process MUST be enclosed within <think> </think> tags. After completing your reasoning, choose only one of the following actions (do not perform both): 

(1) If you find you lack some knowledge, you can call a search engine to get more external information using format: <search> your query </search>. 
(2) If you have enough knowledge to answer the question confidently, provide your final answer within <answer> </answer> tags, without detailed illustrations. For example, <answer>Beijing</answer>.
\end{Verbatim}
\end{tcolorbox}


\section{Ablation Experiments}
\label{app:ablation_omega}
We conduct an ablation study on ALFWorld using the \textbf{Qwen2.5-1.5B-Instruct} backbone to evaluate the impact of the hindsight weighting coefficient $\omega$. This parameter modulates the relative influence of the micro-scale hindsight advantage against the macro-scale GRPO baseline. Specifically, the composite advantage is formulated as defined in Eq. (\ref{eq:hcapo_advantage}).

Here, the first term is the trajectory-level GRPO advantage, while the second term is the step-level hindsight correction; $\omega$ directly scales their relative contribution. By varying $\omega$, we can quantify how strongly hindsight credit assignment affects learning in the Qwen2.5-1.5B setting.

Table~\ref{tab:ablation_omega} reports per-subtask success rates and the overall success rate under four settings ($\omega=0, 0.2, 0.5, 1.0$). The results show a clear monotonic trend: as $\omega$ increases, the overall success rate improves step by step (72.8 $\rightarrow$ 79.7 $\rightarrow$ 84.4 $\rightarrow$ 87.0). In particular, $\omega=0$ corresponds to the GRPO baseline and $\omega=1.0$ reflects our full HCAPO.

This consistent improvement indicates that injecting hindsight credit assignment is highly effective, validating our design and motivating the default choice of $\omega=1.0$.
\begin{table*}[ht]
\centering
\caption{Ablation on ALFWorld with different $\omega$ values. Results are averaged over 3 random seeds.}
\label{tab:ablation_omega}
\resizebox{0.8\textwidth}{!}{
\begin{tabular}{l ccccccc}
\toprule
\textbf{$\omega$} & \textbf{Pick} & \textbf{Look} & \textbf{Clean} & \textbf{Heat} & \textbf{Cool} & \textbf{Pick2} & \textbf{All} \\
\midrule
0   & 85.3$\pm$1.5 & 53.7$\pm$8.0 & 84.5$\pm$6.8 & 78.2$\pm$7.9 & 59.7$\pm$5.0 & 53.5$\pm$5.6 & 72.8$\pm$3.6 \\
0.2 & 89.5$\pm$8.8 & 45.8$\pm$11.8 & 77.8$\pm$5.2 & 87.0$\pm$6.9 & 78.9$\pm$0.0 & 73.6$\pm$8.6 & 79.7$\pm$3.4 \\
0.5 & \textbf{91.4$\pm$6.1} & \textbf{79.2$\pm$5.9} & 93.1$\pm$2.0 & 83.3$\pm$15.6 & \textbf{88.9$\pm$0.0} & \textbf{80.7$\pm$1.0} & 84.4$\pm$4.8 \\
1.0 & 88.6$\pm$7.0 & 75.0$\pm$0.0 & \textbf{97.6$\pm$1.8} & \textbf{90.7$\pm$6.9} & 84.2$\pm$0.0 & 74.2$\pm$6.9 & \textbf{87.0$\pm$4.1} \\
\bottomrule
\end{tabular}
}
\end{table*}

\end{document}